# "Can You Do This?" Self-Assessment Dialogues with Autonomous Robots Before, During, and After a Mission

Tyler Frasca
Tufts University
Boston, Massachusetts, USA
tyler.frasca@tufts.edu

Evan Krause
Tufts University
Boston, Massachusetts, USA
evan.krause@tufts.edu

Ravenna Thielstrom
Tufts University
Boston, Massachusetts, USA
ravenna.thielstrom@tufts.edu

Matthias Scheutz
Tufts University
Boston, Massachusetts, USA
matthias.scheutz@tufts.edu

## ABSTRACT

Autonomous robots with sophisticated capabilities can make it difficult for human instructors to assess its capabilities and proficiencies. Therefore, it is important future robots have the ability to: introspect on their capabilities and assess their task performance. Introspection allows the robot to determine what it can accomplish and self-assessment allows the robot estimate the likelihood it will accomplish at given task. We introduce a general framework for introspection and self-assessment that enables robots to have task and performance-based dialogues before, during, and after a mission. We then realize aspects of the framework in the cognitive robotic DIARC architecture, and finally show a proof-of-concept demonstration on a Nao robot showing its self-assessment capabilities before, during, and after an instructed task.

## 1 INTRODUCTION

Imagine you are tasked to search for wounded people in a natural disaster zone with a team of humans and robots. You know the general capabilities of the robot, for example, that it can search buildings, or avoid obstacles, but you are uncertain whether it will actually make its way through the rubble into a partly collapsed building. You ask, "Robot 7, will you be able to enter the building by parking lot?" "Yes, I have determined that there is an open path I can use," the robot may respond. "Great, go right ahead and search the building for wounded people," you order the machine. As the robot enters the building, you want an update on what it is doing. The robot informs you, "I'm trying to enter a room but the door does not seem to open." "Can you break it open?" you ask. "I don't think I can," the robot replies. "Look for a hole in the wall and if there is one, drive through it," you order. "Roger that," the robot confirms. Eventually, the robot is able to make its way in and reports that there are no people in the room. Upon its return, you ask the robot how it got into the room and it tells you it found an opening in the wall. "Remember to look for such holes in the future," you instruct. The robot stores the new information for future missions.

While these types of task-based interactions are not quite attainable yet, they point to an important capability future robots need to have: introspection into their capabilities and assessment of their actual and predicted task performance. Introspection enables the robot to evaluate what it can accomplish and self-assessment allows the robot to provide estimates of the likelihood it will complete a given task. Self-assessment here can be based on past performance or different ways to extrapolate known performance to unknown cases (e.g., using analogical reasoning [7]).

In this paper, we introduce a framework for introspection and self-assessment that enables robots to have task and performance-based dialogues before, during, and after a mission about the likelihood that they will be able to complete their tasks. The framework is general in that it lays out functional and architectural requirements for self-assessment without prescribing any particular assessment algorithm or architecture, even though the extent to which a robot will be able to assess its performance will depend on the extent to which it can introspect its internal states, track its performance over time, and estimate action success in light of environmental variations. We have implemented the core framework in the cognitive robotic DIARC architecture [4–6] as it provides the natural language capabilities for performance and self-assessment dialogues. We then show a proof-of-concept demonstration on a Nao robot showing its self-assessment capabilities before, during, and after an instructed task.

## 2 INTROSPECTION AND SELF-ASSESSMENT FRAMEWORK

We start by distinguishing three different types of assessment interactions one could have with a robot: before task execution (*a*

*priori*), during task execution (*in situ*), and after task execution (*a posteriori*)[1]. Each of these interactions have different purposes and require the robot to have different introspective capabilities.

*A priori* self-assessment enables dialogues about what a robot can do and when it might do it. The robot could figure, as part of a larger mission planning process, what the robot should do, or in the context of team tasks, what role the robot should assume and its specific goals. The *in situ* interactions take place during task execution and are typically aimed at providing performance updates to the human as well as expected or actual deviations from the planned execution. The third, *a posteriori* assessment, is a retrospective dialogue about what happened, possibly why, and how it could be improved in the future. All three forms of self-assessment can be enabled in the same architecture assuming the necessary introspection mechanisms available which we will discuss next in detail.

## 2.1 A Priori Self-Assessment

When an agent receives an instruction to perform a task, it is desirable for the agent to assess whether it is able to accomplish the task and report back to the human instructor any limitations instead of simply attempting to perform it and possibly failing. Several factors must be taken into consideration for these assessments, for example, what other goals the agent is pursuing or supposed to accomplish, and which of them should take priority. For instance, a new goal to radio headquarters may be of less priority than a current goal of helping someone trapped in a burning building. The agent ought to consider various factors (e.g., task complexity, resources, or time) when assessing which goal to immediately satisfy. To handle this type of assessment, the agent must keep track of all goals it is actively pursuing and must assess tradeoffs among goals.

Once the agent knows what goals to accomplish, it can attempt to determine whether it will be able to perform the associated tasks. This may require it to perform various types of actions (internal on its system and external on the environment). To assess whether an action is applicable in a given context and how likely it will succeed, the agent needs to maintain a set of action pre-conditions, operating conditions, and effects together with associated probabilities. It can then use these probabilities to reason about the likelihood of action success. The agent then needs to repeat the assessment for all actions required to perform each task, possibly making assumptions about various aspects of an action: (1) the likelihood of the pre-conditions being met, including the likelihood that it will get all necessary perceptions right, (2) the likelihood of the operating conditions holding, including the likelihood that it will perform all necessary motor actions right, and (3) the likelihood of the post-conditions holding given its performance of the action and various extraneous circumstances. The way probabilities are combined will depend on the particular action, e.g., whether the actions are performed by the agent alone using only its effectors (e.g., lowering an arm) or whether they require manipulation of an object (e.g., grasping an object) or those of other agents (e.g., handing over an object). Moreover, how the joint task probabilities are calculated will depend on the available information about individual actions (e.g., whether single success probabilities are known or whether the information is distributional) as well as the way tasks are represented (e.g., in terms of mere action sequences of the agent, in which case success probabilities could simply be multiplied, or in terms of conditional action sequences where distributions of over conditions have to be taken into account).

If there are multiple options for achieving goals, the agent must introspect on these options and assess their likely outcomes and tradeoffs, including estimate time for completion, probability of success, or available resources. For example, if the agent has two sequences of actions A and B which allow the agent to save a person in a burning building, then it should base it preference on an assessment of action success: If action A has a mean success probability of 80% and action B of 40%, then A is the better choice, as executing B may result in an unsuccessful retrieval of the trapped person. Similarly, action duration should be assessed in a similar manner when duration, not success is the critical measure: If action A takes two hours to remove fallen structural beams, but action B takes 30 minutes to knock down a door, then it might be better to execute action B to enter a room.

## 2.2 In Situ Self-Assessment

*In situ* self-assessment enables an agent to provide an update on how well it is progressing towards its goals based on the state it currently is in. This measure is a combination of tracking past performance, possibly including deviations from initially expected actions and durations, and estimating expected performance for the remaining tasks to performed in the service of the agent's goals. For the latter all the *a priori* assessment capabilities can be utilized, in particular, in cases where a pre-determined action plan has to be changed and new options need to be evaluated, or when new goals are given to the agent that need to be accommodated (e.g., when a new goal with higher priority like saving a person trapped in a burning building needs to be prioritized and satisfied immediately).

During task execution, the agent needs to continuously observe the environment to ensure it can still execute the planned action sequences required to achieve its goals (e.g., a structural piece of the building might collapse and require the robot to work faster or find an alternative route). Additionally, some actions (e.g., traversing an unstable floor) may require certain environmental states for the action execution to be successful and thus the agent must monitor these states for possible changes throughout execution. Any detected changes could then potentially force the agent to replan by selecting among viable plans based on assessed expected performance and then update its *in situ* performance estimate.

## 2.3 A Posteriori Self-Assessment

After completing a task, *a posteriori* assessment allows the agent to determine in hindsight what options would have been preferable and update its performance estimates. This can enable changes in the agent's task knowledge and self-assessment that will lead to better action choices in the future and more accurate assessments. To enable *a posteriori* assessment, the agent needs to have access to memory traces of what it did and what happened, and calculate actual and possible performance information from those traces. For example, when the agent accomplished the goal in a given context, it needs to update the estimated duration of execution as well as probability of success. Even if the agent did not accomplish the goal, it still needs to update the success probability and ideally assess why it failed and how it can improve its execution. For example, it may explore options based on questions like "why did it take me so long to remove the rubble blocking the door," "did I need to pick up the rocks to move them or could I have pushed them," or "what would have happened if I could not move the rubble." By including observations of the environment in addition to its own performance, the agent can learn more about how its actions affected the world and will thus have a better understanding for when to execute the action in the future and how likely it will succeed.

[1]The categorization is based on the one in Topic 24 of the ONR program announcement #N00014-17-S-F006

## 3 INTEGRATION INTO DIARC

Based on the general introspection and self-assessment framework discussed in Section 2, we extended the cognitive robotic DIARC architecture to enable all three assessment methods through natural language dialogues. DIARC was selected because it already provides both deep introspection capabilities [3] and extensive natural language capabilities [6] (however, the system is not limited to dialogue based assessment, GUI-based interactions are possible as well). DIARC utilizes declarative and procedural knowledge to explicitly represent beliefs and actions, respectively, which allows it to introspect on its knowledge and connect these representations to natural language forms. For the purposes of self-assessment, we extended the action script representation to include action probabilities, and also the Goal Manager component, to improve assessment by enabling more detailed information about action execution.

In order to enable self-assessment dialogues, the robot needs to have access to various knowledge bases, including beliefs, past and current goals, and actions. Hence, we start off by discussing some of the core data representations in DIARC which enable the agent to assess its performance and then discuss how the system leverages these representations for the three types of self-assessment.

### 3.1 Beliefs

To enable assessment, the robot needs to have access to the world state. Additionally, throughout the robot's execution, it will need to evaluate the truth values of certain observable conditions; this "Belief database" generally serves as the transfer point for this type of environmental knowledge between components of the system. The Belief database stores such information in facts (predicate statements of truth) and rules (conditional relationships between statements). Since the state of the world is constantly changing, this database is likewise mutable, and new beliefs can be added or retracted at any time. Queries can be submitted to Belief either to evaluate the truth value of a particular predicate, or to obtain all possible value-bindings to free variables in the predicate that would make it true.

### 3.2 Action Representation

Action scripts are compact ways of specifying hierarchical robot behavior without explicitly modeling the entire state relative to each action. Table: 1 shows an example script for the "Dance" action. They are defined by an action name and associated parameters each with their own given type (e.g., a reference to a graspable object). Each action has a set of pre-conditions that need to be true before the robot can execute the action, operating conditions which must be true throughout execution, otherwise the action will fail, success post-conditions that will be true when execution succeeds, and failure post-conditions that will be true when execution fails. All condition sets are finite, containing first-order formulas over a finite set of predicates. Semantically, pre-conditions determine an equivalence class of world states in a transition system and the operating conditions hold true throughout the execution, the system will end up in a successor state which is a member of the equivalence class of states defined by the success post-conditions; otherwise it will end up in a state in the equivalence class of the failure post-conditions. They contain a finite sequence of action steps (without any additional non-action expressions such as control expressions like "if-then-else" conditions, "for" and "while" loops, event descriptions, observer expressions). Each action step represents another action script, or an action primitive, which is represented in the same manner, except it provides a single operation instead of containing a sequence of steps.

**Action Script**

| | | |
|---|---|---|
| Name: | Dance | |
| Arguments: | agent | |
| Pre-Conditions: | arms(down) | head(straight) |
| Steps: | raise(arms) | lower(arms) |
| | look(left) | look(right) |
| | look(forward) | raise(arms) |
| | lower(arms) | |
| Operating-Conditions: | standing(agent) | |
| Post-Conditions: | danced(agent) | |
| Success Probability: | 0.9 | |

**Table 1: The action script for the dance action.**

We extended the action representation to include success probabilities based on the specific parameterized arguments during execution. Once an action is completed, the probability of the associated parameterized action is updated. Currently, each parameterized action is represented by a single probability, however a distribution or bounds also can be defined.

The DIARC architecture stores these actions in a local database which can be accessed based on name and parameters, or post-conditions. Depending on the situation the agent may need to look up a specific action to execute, e.g. *walk forward* in which case it will query the database for the action "walk" with the parameter "forward". Additionally, the agent may need an action which satisfies a goal, e.g. *holding cup*, and will query an action with a matching post-condition.

### 3.3 Goal Manager

In DIARC, the Goal Manager (GM) is the component responsible for managing agent goals. When the agent receives a goal, specified as a desired state in predicate form, the request is sent to GM, which determines if and when the goal should be pursued, and how to accomplish it. As GM receives goals from other DIARC components (including sub-goals from itself), it first assesses whether the goals are properly formed, and then adds them to an active goal queue. This queue provides a means for the GM to introspect on goals it still needs to satisfy as well as goals that are currently being pursued. To enable introspection of non-active goals, goals that are rejected or have reached a terminal state are removed from the active goal queue and placed in a past goal queue. Each goal has an associated GoalStatus which can change between *pending*, *active*, *suspended canceled*, *satisfied*, or *failed*.

When GM decides that a goal should be pursued (using priority calculations based on utility, cost, etc), the goal is added to an Action Execution Tree. This tree keeps track of all goal executions for both current and past goals, action sequences for each goal, as well as action parameters and ActionStatuses for each action step. This explicit representation is a critical mechanism for introspecting on current and past goals.

As the first step of goal execution, an ActionSelector is used to generate or plan a sequence of one or more actions to execute. When a sequence of actions is found, the selected sequence is added to the Action Execution Tree, where each node corresponds to an action step which has been or will need to be executed. Each node also contains the action name, action parameters, ActionStatus and Justification for why an action step has succeeded or failed. The ActionStatus changes during the stages of execution, and allows GM to easily traverse the Action Execution Tree to find the step(s) being executed or the step(s) that have failed.

When GM is about to execute an action step, it first assesses if the agent can execute the action by checking the action's pre-conditions against the current world state, as well as if it will lead to any forbidden states. If all pre-conditions pass, the system begins to observe the operating conditions to ensure they hold throughout execution. The system then attempts to execute the action, which either executes a primitive action, or in the case of a higher-level action, grows the Action Execution Tree by adding all action steps to the tree. This process is repeated until all actions terminate as primitive actions, or until an action fails. After an action step has been executed, there is a final check to verify that all post-conditions hold. Notice that checking pre-, operating, and post-conditions might also require performing actions, for instance, observing that some world-state has been achieved (e.g., robot is holding an object). These condition checking actions are also stored as part of the Action Execution Tree, making them accessible to the introspection and self-assessment algorithms.

After each action step is executed, the results of the action are used to update the self-assessment model. This might include updating success/failure probabilities, as well as other meta-data relevant to the model (e.g., actions that came before, world state, etc). After all actions have been executed, and goal execution has terminated, the goal is moved from the active goal queue to the past goal queue, but critically the Action Execution Tree for this goal remains intact, enabling a posteriori self-assessment and introspection.

## 3.4 Introspection in DIARC

As we mentioned in Section 2, there are three main task phases when an agent could be required to introspect on its capabilities and goals: *a priori*, *in situ*, and *a posteriori*. Currently, we have incorporated aspects of each into and will use the "Dance" action described in Table: 1 as a running example.

*3.4.1 A Priori Self-Assessment.* Prior to executing an action to satisfy a new goal, for instance to perform the dance action, the agent may want to assess its other active and past goals. Since GM in DIARC maintains active and past goals, it is able to assess which goal to select and try to accomplish. Additionally, we incorporate the ability for the robot to assess if it can accomplish the goal, how it would accomplish the goal, and the probability it would accomplish the goal. In order to assess if it can accomplish the to dance goal, GM queries the action database for a sequence of actions that results in the agent having danced. Currently, if the agent queries the database and assesses that it doesn't know how to accomplish the goal, then it will fail. However, the selection process can be extended to consider exploring the environment, creative problem solving, or asking another agent for assistance. When the agent finds an action, it plans or extracts the steps required to complete the goal. It is then able to evaluate if it is capable of performing each of the action steps, including raising and lowering its arms and looking left and right. To calculate the probability of completing a goal, the system will first check the action representation to see if one is known, which is 0.9 for the dance action. However, say the robot has no experience executing the action or it just learned the action, then it will calculate the probability by the computing the product of the action's step success probabilities. Currently, if this novel action is a primitive action then it will assume a probability of 0.5. After the action is executed, the system updates it the action representation to include the additional experience and if it was completed successfully.

*3.4.2 In Situ Self-Assessment.* In order for a human interlocutor to query a robot about what action it is performing for a particular goal (e.g., what is the current/next/previous step of goal-X), DIARC has been extended with several key pieces of functionality: (1) action traces are explicitly represented in the Action Execution Tree, (2) goals and actions have semantic representation in first-order predicate logic, and (3) the GM allows the Action Execution Tree to be searched by other DIARC components by way of several exposed actions. For example, GM might receive a goal of the form "did(AGENT, getActionDescription(stepOf(LOCATION,GOAL)))". Here, the goal semantics specify that the AGENT should adopt a goal to perform the "getActionDescription" action which takes in the action argument "stepOf(LOCATION,GOAL)". The action argument specifies that an action description should be generated for the LOCATION step of GOAL (e.g., stepOf(current, dance)).

During execution of the "getActionDescription" action, the semantics of GOAL are used to search the active goal queue to find the relevant goal. Once the relevant goal is found, for example the dance goal, the Action Execution Tree for can be searched to the find node (i.e., action step) corresponding to the specified LOCATION, in this case the current step. Then, the semantic representation of the desired action step can be built using the action name, for instance "look", and action parameters, "right". This information is then asserted into the Belief system where it can be used by the natural language understanding system to respond the interlocutor.

One open question about queries of this form is how deep to search into an execution tree. For example, an action to go-to-breakroom might consist of several sub-actions (e.g., go-to-hallway, navigate-to-breakroom, open-door, go-inside), which themselves have several sub-actions. It's unclear if a query about what step is being executed should result in the most low-level sub-action, top-level action, or somewhere in between. The heuristic currently used in DIRAC is to use the top-level action, but more complex goals, such as those that are executed over the coarse of days or weeks will likely need to modulate this heuristic based on several factors (e.g., interlocutor, world state, context, etc).

Additionally, the system is still able to query the probability that an action will succeed. Currently, the system still calculates the probability with respect to all the actions even if it has completed some, thus in the case of the goal to dance it will extract 0.9. The mechanism to extract this information is all there and we can extend this to consider the remaining actions it needs to complete.

*3.4.3 A Posteriori Self-Assessment.* Thus far, DIARC is currently only able to query how the probability of success updates after action execution. However, the framework is set up so it can incorporate additional performance assessments and reasons why it failed.

## 4 DEMONSTRATION

We now demonstrate the introspection and self-assessment capabilities of the framework described in Section 2 which were realized in the DIARC architecture. In this scenario, the robot, a Nao, is asked to assess its knowledge about a goal at three points: *a priori*, *in situ*, and *a posteriori*. A video of the demonstration can be found at: https://youtu.be/gqEqH00JzRI. Note, to demonstrate the Nao's introspection and assessment capability, the instructor has a dialogue with the robot; however, we are not focusing on its capability to provide explanations. Additionally, because the robots movements are relatively short and quick, the instructor asks the robot to pause execution so the instructor has time to query the assessment capabilities. The ability to pause and resume may be useful for online debugging.

The instructor starts off by greeting the Nao.

Human: Hello Dempster.
Robot: Hello Tyler.

### A Priori Assessment

The instructor then, prior to asking the robot to dance, asks it to assess its capability to dance.

Human: Do you know how to dance?
Robot: Yes.
Human: Describe how to dance.
Robot: To dance, I raise my arms, I lower my arms, I look left, I look right, I look forward, I raise my arms, and I lower my arms.
Human: What is the probability that you can dance?
Robot: The probability that I dance is 0.9.

This dialogue segment demonstrates the Nao is able to query its knowledge about what it can do, how it would do it, and the probability the robot would accomplish the goal. In this demonstration, the Nao is configured with a prior probability of 0.9 to dance, however as the robot gains experience it will update this probability.

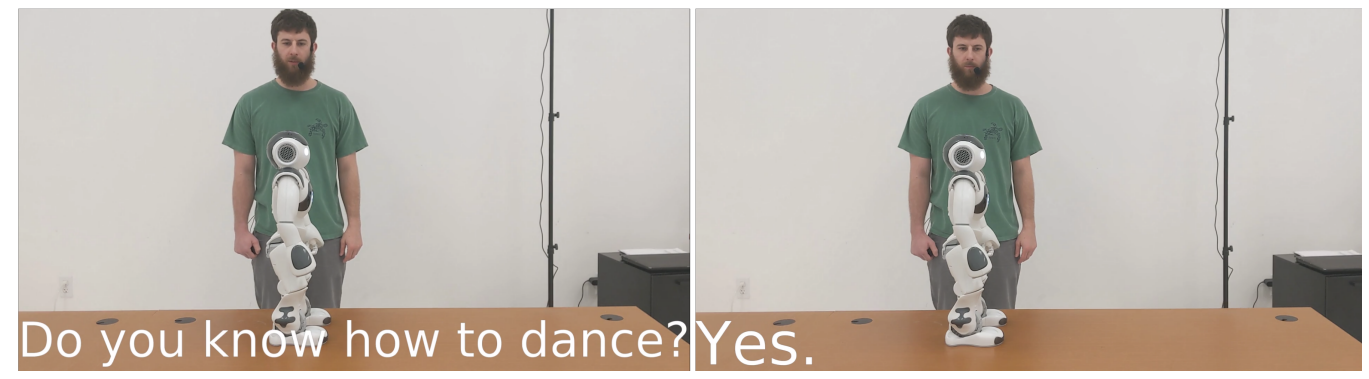

**(a) The Figure above shows the instructor asking the robot if it knows how to dance.**

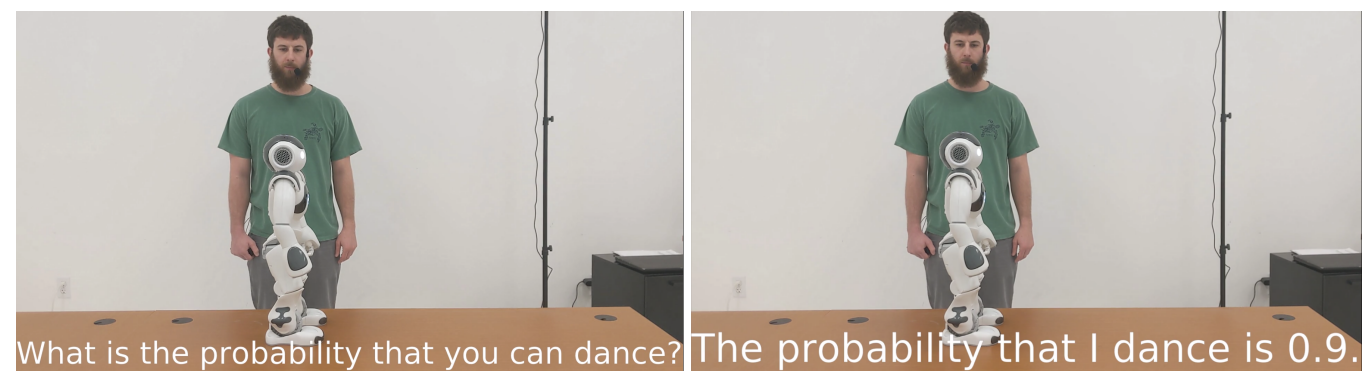

**(b) The Figure above shows the instructor asking the robot to assess the likelihood it will be able to dance.**

### In Situ Assessment

The instructor ask the robot to dance and then to pause execution so it has time to demonstrate the assessment capabilities of the robot *in situ*.

Human: Please dance.
Robot: Okay.
Human: Pause.
Robot: Okay.

Now that the robot has paused, the instructor can query the current progress of the dance action

Human: What is the current step of dance?
Robot: The current step of dance is that I look right.
Human: What is the previous step of dance?
Robot: The previous step of dance is that I look left.
Human: What is the next step of dance?
Robot: The next step of dance is that I look forward.

The robot pauses executing after looking to the left and is able to provide this information to the instructor by using the Action Execution Tree and the ActionStatuses. Once it locates the step in progress, it responds to the instructor.

The instructor tells the robot to continue executing.

Human: Resume.
Robot: Okay.

### A Posteriori Assessment

Finally, after dancing, the instructor asks the robot to assess the updated success probability.

Human: What is the probability that you can dance?
Robot: The probability that I dance is 0.91667.

By comparing the new probability to the original, it is evident the robot successfully updated the probability and is able to introspect on it.

## 5 DISCUSSION AND RELATED WORK

There are currently only a few proposals for robotic self-assessment in the literature (e.g., [2] for navigation, [8] for vision, [1] for human-robot interaction in teams). We have provided a novel proof-of-concept demonstration showing that a robot's introspection capabilities can be used to access different types of task-based knowledge in a systematic fashion, regardless of whether it is before, during, or after task execution. We demonstrated the core functionality of utilizing deep introspection for self-assessment, even though not all introspection capabilities available in DIARC have been utilized (e.g., component-internal introspection [3]) and only one type of self-assessment – probability of task completion – has been implemented, independent of context.

There are two important shortcomings with the current implementation that we will address in future work. Currently the robot needs to have full prior knowledge about the success probabilities of all involved actions, otherwise it cannot compute the overall success probability. However, this information might not be available for all Additionally, it records success probabilities independent of environmental context, but environmental context can have a major impact on action success (e.g., a manipulation action such as "pick-and-place" being performed in an open vs. a cluttered environment). While the agent could either estimate the likelihood of the action completing based on similar actions or use a "default value" that can later be updated, it might be better to determine a rough estimate through repeated *mental simulation* of the action. DIARC has the ability to perform these simulations through duplicating parts of the architecture and connecting it to a physics simulation environment [7]. This type of simulation might also help address the more challenging second point where the robot needs to estimate the success probability of an action under new environmental conditions. Note that in addition to performing actions and learning the probability of them succeeding in different contexts, the agent needs to learn about possible objects and the probability that it will be able to detect them when needed (e.g., [8]). To the extent that the robot has a sufficient model of the environment and can perform mental simulations of the action in the envisioned context, it might be able to generate a sufficiently accurate distribution of success probabilities across multiple simulation runs that can serve as a proxy for the true probability. However, this will impose significant common-sense knowledge requirements on the robot about the objects, including their physical properties and function. Additionally, the robot needs to use models of these objects in the simulated environment that are accurate enough for the robot to perform actions on them that mirror the real world.

## 6 CONCLUSION

Autonomous robots with sophisticated capabilities can make it difficult for human instructors to assess if the robots can accomplish a given task, especially if the humans have not observed them performing the task. This is particularly important in the context of mixed-initiative teams where an unexpected event requires the team to adjust its goals and activities. Being able to query the robot in natural language about its abilities as well as the likelihood of completing a task will go a long way towards making robots more useful team members and building human trust.

We presented a unified self-assessment framework based on deep architectural introspection that enables robots to have dialogues about self-assessment before (*a priori*), during (*in situ*), and after (*a posteriori*) a task. We implemented the framework in the DIARC architecture and demonstrated its core functionality on a fully autonomous Nao robot.

The refinement of the assessment algorithms is ongoing work. This includes the ability to perform simulation-based assessments of actions and tasks the robot has never performed, which will enable the robot to engage in hypothetical and counterfactual dialogues about performance assessment.

## 7 ACKNOWLEDGEMENTS

This work was supported in part by the U.S. Office of Naval Research under Grant #N00014-18-1-2503.